\pdfoutput=1
\documentclass[letterpaper]{article}
\usepackage[totalwidth=480pt, totalheight=680pt]{geometry}
\usepackage{graphicx}
\usepackage{amsmath}
\usepackage{amssymb}
\usepackage{balance}
\usepackage{float}
\usepackage{amsthm}
\usepackage{amsmath}
\theoremstyle{definition}
\newtheorem{example}{Example}
\newtheorem{proposition}{Proposition}
\usepackage{booktabs}

\title{Towards Scalable Network-Based Interventions for HIV Treatment and Prevention via R2C-Agent}
\author{
Akseli Kangaslahti\\
Harvard University\\
\texttt{akselikangaslahti@g.harvard.edu}
\and
Milind Tambe\\
Harvard University\\
\texttt{tambe@seas.harvard.edu}
\and
Alastair van Heerden\\
University of Witwatersrand\\
Wits Health Consortium\\
\texttt{alastair.vanheerden@wits.ac.za}
\and
Cheryl Johnson\\
World Health Organization\\
\texttt{johnsonc@who.int}
}
\date{}

\begin{document}

\maketitle

\begin{abstract}
Human Immunodeficiency Virus (HIV) remains a pressing public health challenge worldwide. Targeting limited HIV prevention and treatment resources effectively with network-based algorithms is a long-standing problem with a rich history at AAMAS and other computational venues. However, optimizing targeted biomedical interventions for minimizing spread of HIV often requires observability of the contact network of HIV transmission risk behavior among individuals, which is prohibitively expensive to collect at scale. Peer referral networks, i.e., networks collected through a recursive peer referral process, are an alternative network class often used to study HIV in hidden populations that can be collected digitally at scale, but they deviate from the underlying contact network: referrals can follow non-risk social ties and participants accept only one referral, hiding contacts among already-recruited individuals. In this work, we first formalize the novel problem of planning network-based interventions to minimize HIV spread in the hidden contact network while observing only the peer referral network. Second, we implement R2C-Agent, a decision making agent for interventions that is designed around our new conditional graph diffusion model called Referral-to-Contact Diffusion (R2C-Diff). R2C-Diff positions the peer referral process as incremental corruption of the underlying contact network and learns to reverse it. R2C-Agent plans its interventions over contact networks sampled from R2C-Diff, so its decisions account for uncertainty about the hidden contact network. Finally, we conduct simulated experiments using real-world HIV network data that demonstrate that R2C-Agent empirically plans more effective interventions than baselines from interdisciplinary literature and highlight key tradeoffs and assumptions driving performance.
\end{abstract}

\section{Introduction} \label{sec:intro}

Deciding which individuals in a network should receive a budgeted intervention to prevent new infections and onward transmission is a long-standing problem at AAMAS and other computational venues. For example, prior work has studied prevention interventions \cite{sambaturu2020designing, khanna2019modeling, brazia2026reconstructing}, prevention information spread \cite{wilder2018end, yadav2016using, yadav2017influence, yadav2018please}, network-based testing \cite{choo2026adaptive, kong2026latent}, and targeted intensive treatment resource distribution \cite{kangaslahti2026network}. In such problem instances, the goal is typically to strategically distribute a budgeted resource amongst a population so as to optimize a key public health outcome, e.g., minimize new infections, maximize new cases found, or maximize preventative information spread. In order to best optimize and evaluate biomedical intervention decisions, such approaches often require observability of the underlying \textit{contact network} of the population, i.e., the network of individuals in which connections capture pairwise transmission risk behavior. However, the contact network for sexually transmitted infections such as Human Immunodeficiency Virus (HIV) is typically prohibitively expensive to collect in practice, especially at scale, creating a substantial barrier to deploying network-based algorithms. 

One tool that can support prevention and treatment efforts is a \textit{peer referral network} \cite{mcknight2006respondent}. These studies use a recursive referral process in which selected seed participants refer peers, who in turn refer others, allowing researchers to collect a network of who referred whom. This approach aims to leverage existing social trust within communities to reach hidden or hard-to-reach populations \cite{heckathorn1997respondent}. In the digital era, these networks can also be easier to scale, as peer referral can be conducted without an in-person visit, e.g., online or through text messages \cite{bengtsson2012implementation, stromdahl2015implementation}. However, the recursive peer referral process introduces deviations from the contact network, e.g., referrals are not restricted to contacts with shared HIV transmission risk behavior and participants only accept one peer referral to join the network (see Figure \ref{fig:deviation}). Thus, though easier to observe at scale, the peer referral network cannot replace the contact network for targeted biomedical interventions that seek to minimize HIV spread.

In this work, we study how to plan budgeted one-shot HIV interventions (in line with one-shot trials and studies \cite{grant2010preexposure, baeten2012antiretroviral, amirkhanian2015effects, kelly2020social}) on a hidden contact network using only the observed peer referral network. To this end, we implement R2C-Agent, a decision making agent that uses our proposed Referral-to-Contact Diffusion (R2C-Diff) model to estimate a distribution over contact networks conditioned on the observed referral network, and then plans its intervention with respect to that distribution (Figure \ref{fig:r2c-agent}). This way, we aim to enable network-based interventions that inherit the scale and reach of peer referral networks with only a small labeled contact network for training (see Section \ref{sec:conclusion}). Our specific contributions are:
\begin{enumerate}
    \item We formalize the problem of planning budgeted network-based interventions for HIV prevention and treatment when the contact network is unobserved and only a peer referral network is available. 
    \item We implement a decision making agent called R2C-Agent (see Figure \ref{fig:r2c-agent}) that is built around our proposed Referral-to-Contact Diffusion (R2C-Diff) model (see Figure \ref{fig:our-method}), a conditional graph diffusion model that positions each peer referral step as an incremental corruption of the contact network and learns to reverse these steps one at a time. R2C-Agent plans its intervention over contact networks sampled from R2C-Diff, so its decisions account for uncertainty about the hidden contact network.
    \item We build an evaluation pipeline on publicly available real-world HIV network data that measures how well an approach supports two complementary interventions, intensive treatment and targeted pre-exposure prophylaxis (PrEP) distribution, across different intervention budgets. R2C-Agent achieves better public health outcomes than baselines from interdisciplinary literature.
    \item We conduct ablation studies that isolate the effects of design choices and assumptions driving intervention quality.
\end{enumerate}

\begin{figure}
    \centering
    \includegraphics[width=0.5\linewidth]{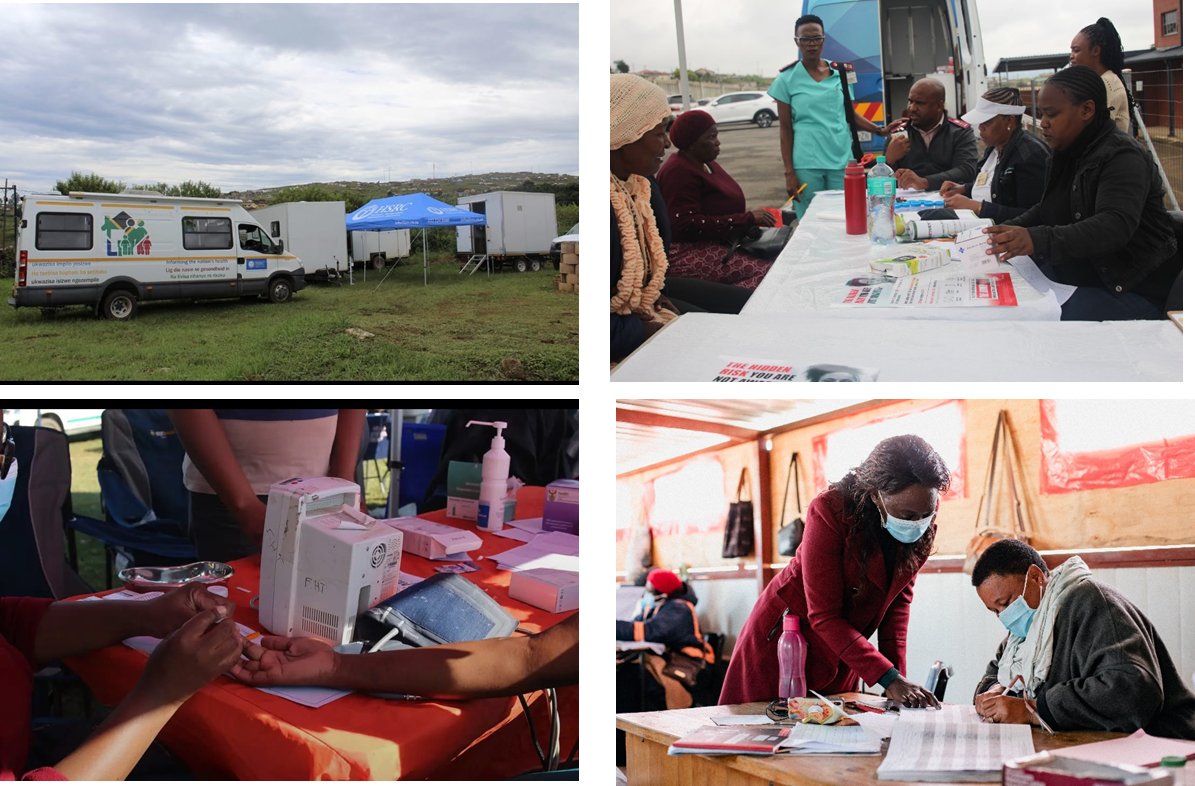}
    \caption{Our collaborators' HIV fieldwork in South Africa.}
    \label{fig:test-distribution}
\end{figure}

We conduct this research in collaboration with domain experts from the World Health Organization and the University of Witwatersrand who are working on a digital peer referral network for providing automated HIV-related resources and support. Figure \ref{fig:test-distribution} shows images of our collaborators' fieldwork.

\section{Background and Related Work} \label{sec:related-work}
\subsection{Human Immunodeficiency Virus (HIV)}
HIV is a sexually transmitted infection that continues to be a pressing public health challenge. As of 2025, more than 40 million individuals are infected worldwide, and about 1.2 million new infections are still occurring annually \cite{unaids2026}. Though there is currently no cure available for HIV, proper treatment via Antiretroviral Therapy (ART) can help people living with HIV reach viral suppression, after which they effectively can no longer transmit the virus and progression to Acquired Immunodeficiency Syndrome (AIDS) can be stopped. As such, UNAIDS has set the 95-95-95 targets for HIV \cite{unaids95}, which aim for 95\% of individuals living with HIV to know their status, 95\% of individuals that are aware of their HIV-positive status to receive treatment, and 95\% of individuals that receive treatment to achieve viral suppression. Although some estimates indicate important progress towards these goals, e.g., in 2025 AIDS-related deaths reached their lowest level in 30 years \cite{unaids2026}, many others remain less optimistic. For example, ART coverage sits at only about 78\%, and UNAIDS has stated the world is not on track for the ultimate 2030 goal of ending AIDS as a public health threat \cite{unaids2026}. Efficient network-based interventions are an especially pressing topic to study as public health agencies that are already constrained are currently facing additional budget cuts for HIV prevention \cite{ten2025impact}.

\subsection{Network-Based Interventions} \label{sec:interventions}
Several types of network-based interventions for HIV treatment and prevention have previously been studied. Prior work has analyzed targeting strategies for pre-exposure prophylaxis (PrEP), a preventative medication that is used to protect individuals who do not have HIV but are at risk of infection \cite{roberts2022impact}. Analogous work on modeling and vaccinating against epidemics in general has also been presented at AAMAS, e.g., \cite{xia2014comparison, singh2018behavior, sambaturu2020designing}. Additional AAMAS papers involve selecting peer leaders for maximizing spread of HIV-related information \cite{yadav2016using, yadav2017influence, yadav2018please, wilder2018end}, though these decisions are made with respect to the social/influence network rather than the contact network. Furthermore, other intervention efforts have proposed algorithms for distributing limited intensive treatment resources, e.g., adherence counseling or second- or third-line ART, to individuals living with HIV whose first-line ART is failing \cite{kangaslahti2026network}. Efficient network-based testing algorithms have also been studied to maximize casefinding with limited testing resources \cite{choo2026adaptive}. Some of these HIV intervention works assume observability of the contact network \cite{kangaslahti2026network, kong2026latent}. Others accept the lack of observability in the contact network and propose heuristics for interventions, such as highest-degree selection \cite{liu2021efficient, brazia2026reconstructing} or learned predictions of individual risk scores that disregard network impacts \cite{marcus2019use, haas2021machine}. Influence maximization efforts at AAMAS sometimes query or sample the unobservable network to learn information about it before making a decision \cite{yan2019maximization, kamarthi2019influence}. Some papers have also introduced methods for predicting unseen HIV-related networks from data that is easier to observe. One such paper infers the full underlying social network from RDS data with evaluation on prediction accuracy \cite{chen2016seeing}. Perhaps the most similar work to ours infers the contact network based on egocentric data and/or sociodemographic and injection-venue features, then studies how well the predicted networks can forecast the impact of fixed interventions compared to ground truth \cite{clipman2026geometric} or support heuristic intervention strategies without comparing to ground truth \cite{brazia2026reconstructing}. In contrast, we use the peer referral network as the condition for inferring the contact network and evaluate the quality of individual-level targeting decisions that sampled networks support, measured on the ground truth contact network. 

\subsection{Relevant Graph Generation} \label{sec:cond-diff}
Existing graph diffusion work is often focused on unconditional generation that produces graphs from scratch, e.g. \cite{vignac2022digress, xu2024discrete}. Some papers condition generation on graph-level properties, such as Graph DiT for molecule graph generation with desired molecular properties \cite{liu2024graph}. Others condition on node-level information, e.g., EDGE proposes a graph generation method that can be guided by node degrees and frames corruption as incremental edge deletion \cite{chen2023efficient}. The closest work for our task is graph-to-graph generation \cite{jin2018learning}, though this is intended for finding molecules with better properties than the input molecule, and the authors use adversarial learning rather than graph diffusion. Another related learning task is graph completion and inpainting via models like DiffLinker \cite{igashov2024equivariant}, where the goal is to fill in missing parts of a known subgraph. Methodologically, the closest models to R2C-Diff are autoregressive graph diffusion models like GraphARM \cite{kong2023autoregressive} that sample one node at a time, though they are often intended for unconditional generation and attempt to learn a good node ordering, whereas node ordering is part of the data in our case. Network prediction and inference problems also have a history at AAMAS, e.g., adversarial link prediction problems \cite{godziszewski2023adversarial} and agent coordination graph inference \cite{li2020deep}. In general, most graph generation is unconditional, trained on at least thousands of nodes of data (here we have less than 700), and evaluated using graph similarity statistics rather than downstream intervention/decision quality, making most prior models unsuitable for our learning task.


\section{Problem Formulation} \label{sec:formulation}
\subsection{Notation}
We consider a fixed population of people $V$, which we refer to as nodes. Nodes have contacts and social edges with one another, decide whom to refer during peer referral, and engage in simulated HIV risk behavior with their contacts during evaluation.
$V$ is partitioned into \emph{positive nodes} $V^{+}$ (people living with HIV) and \emph{negative nodes} $V^{-}$ (people without HIV), and each node $v$ has a covariate vector $x_v$.
All networks below are undirected graphs on $V$.
\begin{enumerate}
    \item \textbf{Contact network} $G_c = (V, E_c)$: edges are pairwise HIV transmission-risk ties (e.g., sexual or injection-sharing contacts). This is the hidden ground truth on which HIV spreads.
    \item \textbf{Social network} $G_s = (V, E_s)$: social ties without transmission risk, with $E_s \cap E_c = \emptyset$.
    \item \textbf{Interaction network} $G = (V, E)$ with $E = E_c \cup E_s$: the joint social and contact network.
    \item \textbf{Referral network} $G_r = (V, E_r)$, $E_r \subseteq E$: the observed network of who referred whom. Each node accepts at most one referral, so $G_r$ is a forest. The nudge order $\pi$ lists nodes in the order in which they were ``nudged'' to refer peers.
    \item \textbf{Predicted contact networks} $\hat{G}_c^{(1)}, \ldots, \hat{G}_c^{(k)}$: samples from $p_\theta(\cdot \mid G_r, \pi, V^{+}, V^{-},x)$, a model of the contact network given the observations.
\end{enumerate}

\subsection{Formulation}
\paragraph{Peer referral process.} We consider a class of peer referral network studies that are conducted as follows:
 \begin{enumerate}
     \item Seed nodes are recruited by researchers. They are surveyed and/or offered HIV testing, revealing their HIV status and individual covariates.
     \item Seed nodes receive ``nudges'' from the researchers, asking them to recruit their peers to join the network.
     \item Seed nodes refer their peers, who recursively repeat this process until no recruited node remains to be nudged, i.e., the frontier is empty.
 \end{enumerate}
 
 \paragraph{Observations.} In this scenario, we can observe $\mathcal{O} = (G_r, \pi, V^{+}, V^{-}, x)$: the peer referral network $G_r$, the nudge order $\pi$, the partition $V = V^{+} \cup V^{-}$, and the covariates $x_v$ for every $v \in V$. The contact network $G_c$ and social network $G_s$ are unobserved. We know that every referral edge is a contact or social tie ($E_r \subseteq E_c \cup E_s$) but we never observe which one. We also assume that peer referral seeds are selected from different communities, i.e., that the connected components of $G_r$ coincide with those of $G$.

\begin{figure}
     \centering
     \includegraphics[width=0.8\linewidth]{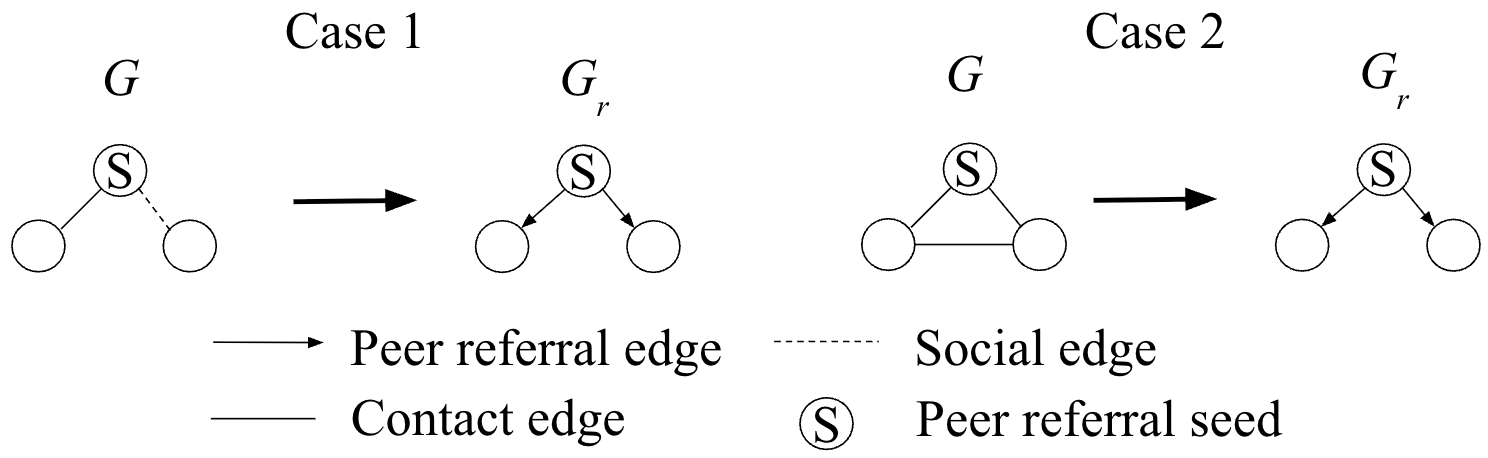}
     \caption{Examples of deviation between the hidden contact network $G_c$ and the observable peer referral network $G_r$. $G_c$ is a subgraph of the hidden joint contact and social network $G$ through which peer referrals are conducted. In Case 1, a social edge from $E_s$ appears in $G_r$ but not in $G_c$. In Case 2, the contact edge between the two recruits of the seed node S is never recorded in $G_r$ because recruits can only accept one referral to join $G_r$.}
     \label{fig:deviation}
 \end{figure}

 There are two deviations from $G_c$ that are introduced by the peer referral process, which we illustrate with examples in Figure \ref{fig:deviation}. First, nodes recruit along both contact and social ties, so some referral edges carry no HIV transmission risk. Second, nodes only accept one referral to join the network, so $G_r$ is a spanning forest of $G$ and contacts among already-recruited nodes are never recorded.

 \paragraph{Spread and interventions.} Disease spreads from positive nodes $V^{+}$ to negative nodes $V^{-}$ over $G_c$ according to a stochastic spread model. An intervention is a set of nodes $S \subseteq V_{\text{elig}}$ with $|S| \leq B$, where $B$ is the budget and $V_{\text{elig}}$ is the set of nodes eligible to receive it (e.g., $V^{-}$ for HIV PrEP distribution). Applying $S$ changes how members of $S$ participate in spread. We write $\sigma(S; G_c)$ for the expected number of new infections when $S$ is applied, where the expectation is over the randomness of the spread model.

 \paragraph{Objective.} A decision making agent must choose an  intervention that minimizes expected new infections in the hidden contact network. Formally, an agent is a policy $A$ that maps the observations $\mathcal{O} = (G_r, \pi, V^{+}, V^{-}, x)$ and budget $B$ to an intervention $S = A(\mathcal{O}, B)$ with $S \subseteq V_{\text{elig}}$ and $|S| \leq B$. We seek to find:
 \begin{equation}
     \min_{A} \; \sigma(A(\mathcal{O}, B); G_c)
 \end{equation}
 Since $G_c$ is unobserved, $\sigma(S; G_c)$ cannot be evaluated when choosing $S$. The agent must act under uncertainty about $G_c$.

\paragraph{R2C-Agent}
 When designing our decision making agent, we must navigate two key challenges:
 \begin{enumerate}
     \item \textbf{Ambiguity (C1)}: A single referral network is consistent with many different contact networks. Furthermore, it is unclear how the outcome of an intervention changes across these different possible contact networks. Thus, the decision making agent must account for uncertainty over $G_c$.
     \item \textbf{Scarce Supervision (C2)}: We assume only a relatively small labeled dataset with both $G_c$ and $G_r$ collected on the same population. Thus, the agent must learn from scarce data.
 \end{enumerate}

\begin{figure}
     \centering
     \includegraphics[width=0.8\linewidth]{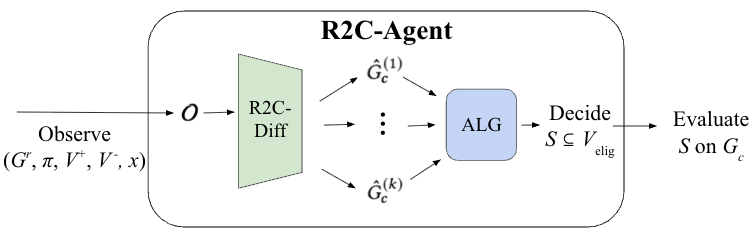}
     \caption{R2C-Agent overview. R2C-Agent uses R2C-Diff to sample possible contact networks conditioned on the observations, then plugs them into an intervention algorithm (e.g., CAST \cite{kangaslahti2026network} or SandIMIN \cite{wang2024efficient}) represented abstractly as ALG. R2C-Agent then outputs ALG's aggregated decision $S \subseteq V_{\text{elig}}$, which respects uncertainty in the hidden contact network. }
     \label{fig:r2c-agent}
 \end{figure}

 We address these challenges with R2C-Agent, a decision making agent built around our R2C-Diff conditional graph diffusion model. R2C-Agent uses sampled contact networks drawn from R2C-Diff in combination with intervention algorithms that assume observability of the contact network to make decisions, as illustrated in Figure \ref{fig:r2c-agent}. We will discuss in the following section how R2C-Agent addresses C1 and C2.

\section{Methods} \label{sec:methods}
\subsection{Generative Modeling and Decision Making} \label{sec:generative}
R2C-Agent couples a generative model of the contact network with an existing planner for known networks. Given the observations $\mathcal{O}$, it draws $k$ contact networks $\hat{G}_c^{(1)}, \ldots, \hat{G}_c^{(k)} \sim p_\theta(\cdot \mid \mathcal{O})$ and passes all of them to the planner. It then produces a decision over these aggregated samples, rather than committing to one contact network with greatest likelihood (the mode). This helps our agent make decisions with respect to uncertainty in $G_c$, addressing C1 (ambiguity). We demonstrate the advantage of this approach in Example \ref{example:mle}, which is analogous to Proposition 1 from \cite{kangaslahti2026policy}.

\begin{example} \label{example:mle}
    Suppose that, given the observed $\mathcal{O}$, there are only two possible contact networks: $G_1$ and $G_2$ with probabilities $P(G_c = G_1 \mid \mathcal{O}) = 0.6$ and $P(G_c = G_2 \mid \mathcal{O}) = 0.4$. Next, suppose that there are only two possible abstract actions in the downstream intervention problem, $a_1$ and $a_2$. Let $U(a \mid H)$ be a function that measures the utility of action $a$ on a contact network $H$, with higher utility being more desirable. Now suppose that we have $U(a_1 \mid G_1) = 5$, $U(a_2 \mid G_1) = 1$ and $U(a_1 \mid G_2) = 1$, $U(a_2 \mid G_2) = 10$. Note that $\mathbb{E}_{G_c}[U(a_1|G_c)] = 0.6 \cdot 5 + 0.4 \cdot 1 = 3.4$, while $\mathbb{E}_{G_c}[U(a_2|G_c)] = 0.6 \cdot 1 + 0.4 \cdot 10 = 4.6$. 
    
    Assume perfect learning. If we learn to predict the mode $G_1$ and pass it to an optimal intervention algorithm as the ground truth, we will take action $a_1$. If we instead learn a generative model, we can sample the distribution of possible contact networks and aggregate action utilities across samples to estimate the expected utilities of $a_1$ and $a_2$. With enough samples, we will select $a_2$, which is the optimal action in expectation.
\end{example}

We extend this argument by noting that this advantage is specific to decision making. If we were instead interested in reconstructing the network as accurately as possible, then a single best estimate would be preferable since samples are less interpretable and less accurate in expectation:

\begin{proposition} \label{prop:sample-accuracy}
Let $G_c \sim p(\cdot \mid \mathcal{O})$, and for each pair of nodes $e$ let $p_e = P(e \in E_c \mid \mathcal{O})$ be its marginal probability of being a contact edge. For any fixed graph $\hat{G}$, the expected edit distance to $G_c$ is at least $\sum_e \min(p_e, 1 - p_e)$, with equality when $\hat{G}$ is the expected edit distance minimizer, i.e., the graph that contains only the pairs with $p_e \geq 1/2$. A sample $\hat{G} \sim p(\cdot \mid \mathcal{O})$ drawn independently of $G_c$ given $\mathcal{O}$ has expected edit distance $\sum_e 2 p_e (1 - p_e)$, which lies between 1 and 2 times this minimum.
\end{proposition}
\begin{proof}
The edit distance counts the pairs on which $\hat{G}$ and $G_c$ disagree. The probability of disagreement on a single pair $e$ depends only on the marginals of $e$. If $\hat{G}$ is fixed, the disagreement probability is $1 - p_e$ when $e \in \hat{E}$ and $p_e$ otherwise, so it is at least $\min(p_e, 1 - p_e)$, with equality for the expected edit distance minimizer. Thus, by linearity of expectation, the expected edit distance is at least $\sum_e \min(p_e, 1-p_e)$, with equality at the expected edit distance minimizer.

If $\hat{G}$ is an independent sample from $p(\cdot \mid \mathcal{O})$, the probability of disagreement with $G_c$ on $e$ is $p_e(1 - p_e) + (1 - p_e)p_e = 2p_e(1 - p_e)$. Since $2p_e(1 - p_e) = 2\min(p_e, 1-p_e)\max(p_e, 1-p_e)$ and $\frac{1}{2} \leq \max(p_e, 1 - p_e) \leq 1$, we have $\min(p_e, 1 - p_e) \le 2p_e(1 - p_e) \le 2\min(p_e, 1 - p_e)$. By linearity of expectation, the expected edit distance of $\hat{G}$ is $\sum_e 2 p_e (1 - p_e)$, which lies between 1 and 2 times the minimum of $\sum_e \min(p_e, 1-p_e)$.
\end{proof}
Proposition~\ref{prop:sample-accuracy} implies that even a perfectly learned generative model can appear up to twice as inaccurate as the best single reconstruction, while still supporting better decisions. Accordingly, we evaluate decision making agents primarily by the outcomes of the interventions they support, and report reconstruction accuracy as a secondary metric (Section \ref{sec:evaluation}).


\subsection{R2C-Diff} \label{sec:ours}
\begin{figure}
    \centering
    \includegraphics[width=0.75\linewidth]{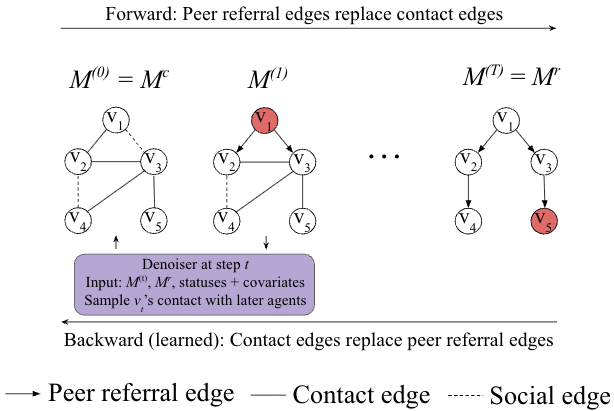}
    \caption{Overview of R2C-Diff. In the forward process, R2C-Diff starts with the true contact network with adjacency matrix $M^{(0)} = M^c$, then incrementally plays out the recorded peer referral process until it reaches the observed peer referral network with adjacency matrix $M^{(T)} = M^r$. The denoiser learns to reverse this process, step by step.}
    \label{fig:our-method}
\end{figure}

To support our decision making agent, we propose Referral-to-Contact Diffusion (R2C-Diff), a conditional graph diffusion model that learns an approximation $p_{\theta}(G_c \mid \mathcal{O})$. The core idea is to treat each peer referral step as a known, incremental corruption of the clean contact network, then learn to reverse this process. Figure \ref{fig:our-method} shows an overview of R2C-Diff, which is described further in the following paragraphs.

Training R2C-Diff requires a small amount of labeled data where both the referral network and the true contact network are observed (see Section \ref{sec:conclusion}). The trained model and R2C-Agent can then be applied to larger-scale populations where only the referral network is available. As a generative model, R2C-Diff represents uncertainty over contact networks for the agent to plan over (see Section \ref{sec:generative}), addressing C1 (ambiguity). It also leverages the known structure of corruption to decompose the generation into small conditional problems, making efficient use of training data and simplifying what the model actually needs to learn, addressing C2 (scarce supervision; see next paragraphs).

\paragraph{Forward Process.}
Since the connected components of $G_{r}$ and $G$ (but not necessarily $G_c$) are equivalent, we can process each connected component separately without considering edges between separate components. Consider a component whose nodes, in nudge order observed through $\pi$, are $v_1, \ldots, v_T$ and let $M^{c}$ and $M^{r}$ be its contact and referral adjacency matrices in this order. The forward process starts from the clean state $M^{(0)} = M^{c}$. At step $t$, node $v_t$ is nudged, and its edges to all later nodes $\{v_u:u > t\}$ are replaced by its referral edges to them, and all other pairs in the graph remain unchanged. Each pair $(v_i, v_j)$ is thus replaced exactly once, at step $\min(i, j)$, so the state after $t$ steps has the form:

\begin{equation} \label{eq:forward}
    M^{(t)}_{ij} =
    \begin{cases}
        M^{r}_{ij} & \text{if } \min(i, j) \le t, \\
        M^{c}_{ij} & \text{otherwise},
    \end{cases}
\end{equation}

At the end of the forward process, we have $M^{(T)} = M^{r}$, the observed referral network. Note that $T$ is equal to the size of the component, meaning it can vary across components. This corruption process is what sets R2C-Diff apart from prior work. In standard graph diffusion models, artificial noise is applied independently to every pair \cite{austin2021structured, vignac2022digress}, so a conditional variant needs to learn how the referral graph relates to the contact graph from scratch without seeing the peer referral mechanism or the nudge order that generated it. Here, we build the mechanism into the forward process and give the model the nudging order directly, simplifying the scope of the prediction that it needs to make at each step of denoising to help make efficient use of data and mitigate C2 (scarce supervision).

\paragraph{Denoiser.}
To reverse step $t$, the denoiser must predict from the state $M^{(t)}$ the probability that $v_t$ shares a contact edge with each later node $v_u: t < u \leq T$. We use a well-established Graph Neural Network (GNN) architecture with message-passing \cite{gilmer2017neural} that maintains dense node and pair embeddings. Node embeddings are initialized with HIV status and an indicator of whether the node is $v_t$, which informs the model which edges are currently being decided. Pair embeddings are initialized from two edge channels. The first is the current state $M^{(t)}_{ij}$, which is updated as the model samples edges, and the second is the peer referral edges $M^{r}_{ij}$, which stays fixed so that the model can stay grounded with the observed peer referral network even after it gets overwritten in the current state. Node covariates can also be added either as part of the node embeddings, as additional pairwise channels, or both. Over three rounds of message passing, each pair's edge embedding is updated using the embeddings of its two endpoints, and each node's embedding is updated by aggregating messages weighted by its incident edge embeddings, so final embeddings capture the node's three-hop neighborhood. This helps the model consider neighboring nodes and edges as context for its decisions. A pairwise readout then maps each candidate pair's individual node embeddings and their shared pair embedding to a single logit. The model's parameter count is deliberately low to limit overfitting, given data constraints (C2).

\paragraph{Training.}
Each training example covers a single step of denoising for a single component in the training data. The input is always the state $M^{(t)}$, $M^{r}$, the node features, and pair channels, and the target is always $M^{c}_{t, u}$ for all nodes $v_u:t < u \leq T$. We use teacher forcing during training, i.e., inputs use the true intermediate states, which can be computed using Equation \ref{eq:forward}, rather than the model's own samples. This allows for stable supervised training, at the cost of possible error accumulation at inference. We train using binary cross-entropy (BCE) loss over $v_t$'s candidate pairs.

Note that the structure of R2C-Diff allows us to extract one supervised example per node per component, with each example covering just a small conditional problem.

\paragraph{Inference.}
At inference, we start with $M^{(T)} = M^r$, i.e., the peer referral network, which we have directly observed. For $t = T$ down to 1, we repeat the following:
\begin{enumerate}
    \item Run the denoiser on the current state to get $v_t$'s contact probabilities for each node $v_u: t < u \leq T$.
    \item Sample each candidate edge as an independent Bernoulli draw.
    \item Write the sampled row into the state, replacing $v_t$'s referral edges with its sampled contacts. This produces $M^{(t-1)}$.
\end{enumerate}
The output after the last step is one sampled contact network for the given component. Combining the sampled components gives the full sampled contact network. R2C-Agent samples $k$ networks $\hat{G}_c^{(1)}, \ldots, \hat{G}_c^{(k)}$ and feeds them to the intervention algorithm. With one denoiser call per node in the network and $k$ sampled networks, the total number of inference calls is $O(|V|\cdot k)$. Components are independent, so inference can be parallelized across components.

The tradeoff of R2C-Diff is that, unlike in traditional diffusion models, each node's edges are considered and sampled exactly once, so the model is committing to its decisions as it goes. However, every later denoising step's input includes the edges that the model committed to so far, so it conditions its decisions on its past choices.

\balance


\section{Experiments and Results}\label{sec:experiments}
See Appendix \ref{sec:implementation} for additional implementation details regarding R2C-Diff, data, and experiments. All code is in the code appendix and will be made public via GitHub in the camera-ready paper.
\subsection{Dataset} \label{sec:data}
Given our focus on practical, real-world deployment for applied network interventions, evaluating on authentic human social and epidemiological structures rather than synthetic graph generators is essential. However, real-world paired datasets capturing both ground-truth HIV contact networks and observed peer referral networks for the exact same population are exceptionally rare. We therefore base our evaluation on one comprehensive real-world dataset provided by the Inter-University Consortium for Political and Social Research (ICPSR) \cite{ICPSR}, which is anonymized and publicly available upon agreeing to terms and conditions. Our evaluation spans multiple distinct connected components and incorporates many stochastically generated peer referral networks in testing to account for variance in seed node selection and referral trajectories. 

The ICPSR dataset contains a real-world joint social and contact network, i.e., $G$, along with individual-level covariates and infection status for each person in the dataset. We extract an HIV network from this dataset for our experiments, which has 1386 nodes, of which 248 have HIV, and 3793 edges, of which 1381 are contact edges. We only extract gender and sexual behavior-related covariates and incorporate them as two pair channels that encode whether the two nodes agree or disagree on each covariate. Some individuals do not have anything recorded for these covariates, which gets encoded as its own category in the pair channels. The contact network can be easily extracted from the data by filtering for contact edges. Since this is a new direction for network-based interventions, no publicly available dataset that we are aware of collects both a peer referral network and a contact network on the same population. As such, we extract a peer referral network from the ICPSR data by simulating a peer referral process:

\begin{enumerate}
    \item We randomly select a seed node from each connected component in the joint social and contact network. Upon joining the network, the node reveals their HIV status and covariates.
    \item Each seed node receives a nudge, prompting them to refer their peers along all of their contact and social edges without disclosing edge types. 
    \item We recursively repeat this process until all nodes have been recruited to the network, leaving us with the forest-structured $G_r$. Nodes from the recruitment frontier are selected for nudging in random order. We always record all referrals from a nudge before sending out the next nudge.
\end{enumerate}

This simulation aims to mirror real peer referral processes, but the key assumption is that nodes recruit along of their edges. We evaluate the effect of this assumption in our experiments. Since randomness in seed selection and nudging order yields many possible peer referral networks for one contact network, we sample 20 peer referral networks from each component in $G$ for training.

\subsection{Evaluation Pipeline} \label{sec:evaluation}
We evaluate approaches on the quality of the interventions they produce with two complementary one-shot limited resource allocation tasks. In both, the goal is to minimize the expected cascade of new HIV infections in the network after the budgeted intervention.

\textbf{Intensive Treatment}: Here, we are distributing a limited amount of intensive treatment resources, e.g., adherence counseling, second- or third- line ART, to combat first-line ART failures amongst people with HIV, so $V_{\text{elig}} = V^{+}$. We use the CAST algorithm \cite{kangaslahti2026network} to make this one-shot decision. As in prior work \cite{kangaslahti2026network}, we assume that all nodes in $V^{+}$ are experiencing first-line ART failure for the sake of these experiments and that all nodes that are selected for treatment reach viral suppression and therefore do not transmit HIV further. 

\textbf{Targeted PrEP}: The second intervention is the \textit{Targeted PrEP} problem. Here, we can distribute a limited amount of HIV PrEP to nodes in the network that do not have HIV in order to protect them from acquiring HIV, so $V_{\text{elig}} = V^{-}$. We assume that a node receiving PrEP is fully protected, i.e., that it can neither be infected nor pass infection on. We use the SandIMIN algorithm for this decision \cite{wang2024efficient}.

Both the CAST and SandIMIN algorithms have proven performance guarantees and take a budget and a contact network that they assume to be the ground truth as input. In our experiments, R2C-Agent queries R2C-Diff for sampled contact networks and plugs them into the CAST and SandIMIN algorithms, then outputs their aggregated decision. Then, we evaluate the resulting decisions on average new infections in the ground truth contact network. The algorithms and the evaluation assume the same probabilistic spread model, known as the Independent Cascade (IC) model \cite{kempe2003maximizing}. Under this model, infection spreads in a cascade from HIV-positive to HIV-negative nodes, each contact transmitting independently with a fixed probability. This allows for Monte Carlo sampling for spread estimation, both internally in the CAST and SandIMIN algorithms and again (independently) in the evaluation. As such, we can naturally integrate sampled contact networks into the CAST and SandIMIN algorithms: R2C-Agent samples $k = 50$ contact networks and allows the CAST/SandIMIN algorithms to draw one Monte Carlo sample from each in order to estimate spread. For single-prediction baselines (see Section \ref{sec:baselines}), we allow the CAST/SandIMIN algorithms to draw 50 Monte Carlo samples from the single predicted contact network in order to estimate spread, keeping the internal Monte Carlo sampling budget for algorithms constant.

\paragraph{Aggregated Evaluation.}
In our experiments, we partition the HIV network data into two halves with similar distributions of connected component size, preserving connected components across the partition. We then evaluate on two complementary train-test splits in which each half of the data serves as training data for one split and test data for the other. Since our dataset has 1386 nodes total, we are training on about 693 nodes of data during each split. We draw 10 sampled peer referral networks from each test split to account for randomness in the peer referral simulation. We pass these as inputs to each approach, and evaluate each decision across the 10 sampled peer referral networks on their shared true contact network. Each decision is evaluated using 100 averaged Monte Carlo samples of HIV spread on the true contact network. Results are pooled across the 10 sampled peer referral networks and then across the two splits. In summary, each budget point that we record for each approach is averaged over 2 fold $\times$ 10 peer referral network draws $\times$ 100 Monte Carlo samples of spread.

\subsection{Baselines} \label{sec:baselines}
We evaluate several baselines where we swap out different parts of R2C-Agent. These approaches are drawn from interdisciplinary literature and used for both intervention problems:
\begin{enumerate}
    \item \textbf{No Intervention}: The decision making agent decides $S = \emptyset$. This helps measure the effect of the intervention.
    \item \textbf{Random}: The decision making agent selects a random subset of $V_{\text{elig}}$ of size $B$ to receive resources. It does not use the CAST or SandIMIN algorithms. This helps measure the effect of an untargeted intervention.
    \item \textbf{Highest Degree}: The decision making agent selects nodes from $V_{\text{elig}}$ that have the highest degree in the peer referral network to receive resources. It does not use the CAST or SandIMIN algorithms. This is a popular heuristic that has been studied in prior public health literature \cite{liu2021efficient, brazia2026reconstructing}.
    \item \textbf{Pairwise Risk Classification}: The decision making agent trains a classifier to predict the probability that a contact edge exists between two nodes based on their status and features. It uses an L2-regularized logistic regression model with calibrated thresholding to make one independent prediction per pair, resulting in a single predicted contact network. Then, it plugs this network into the CAST and SandIMIN algorithms to produce intervention decisions. This is analogous to individual risk scores that are common in public health literature \cite{haas2021machine, marcus2019use}, applied to edges instead of nodes.
    \item \textbf{Identity}: The decision making agent directly uses the observed peer referral network as the true contact network, and feeds it to the CAST and SandIMIN algorithms for decisions.
    \item \textbf{Omniscient}: This baseline observes the true contact network and directly plugs it into the CAST and SandIMIN algorithms to produce decisions. This serves as an infeasible baseline on average new infections that helps measure the regret of not collecting the contact network.
\end{enumerate}

\subsection{Results} \label{sec:results}
Our core experiment measures the intervention quality of R2C-Agent compared with baselines on the Intensive Treatment and Targeted PrEP problems by average new infections following the intervention across intervention budgets from 0 to 1. 

\begin{figure}[t]
    \centering
    \includegraphics[width=0.5\linewidth]{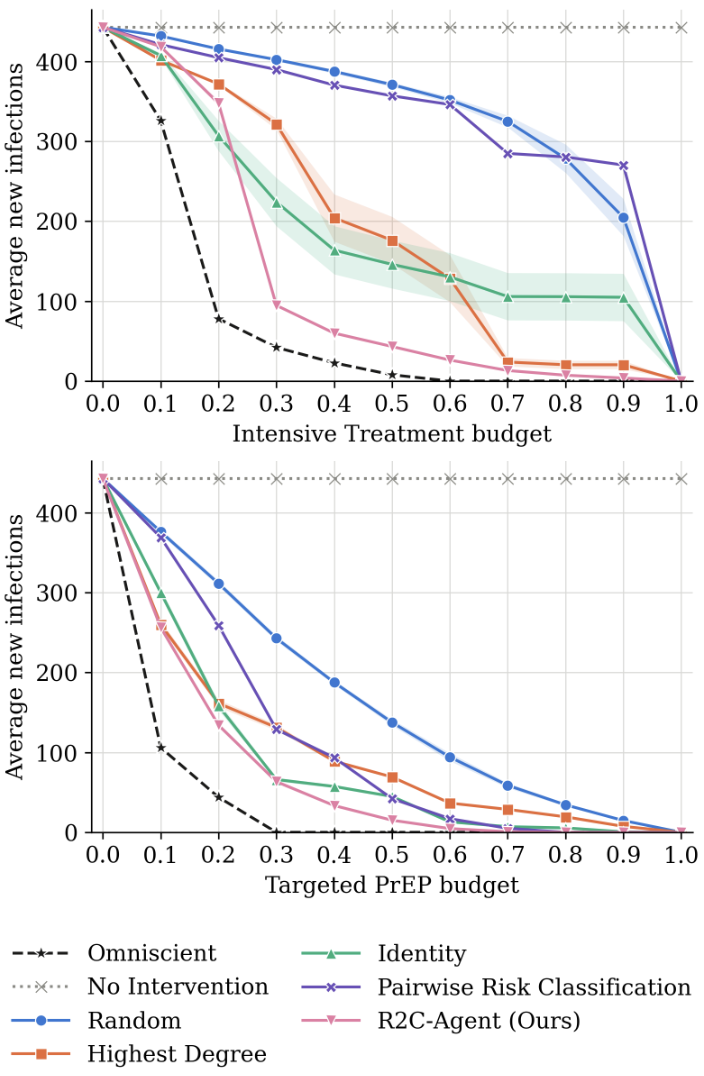}
    \caption{Average new infections $\pm$ stratified standard error (SE) across pooled results on both folds for different intervention budgets (as a fraction of $|V_{\text{elig}}|$) using different decision making strategies. We evaluate on the Intensive Treatment (top) and Targeted PrEP (bottom) problems. Stratified SE is very small compared to the plotted range.}
    \label{fig:core-experiment}
\end{figure}

We aggregate results of each approach using the \textit{Total Regret} of not collecting the contact network, i.e., the difference between the average number of new infections under a given approach and the average number of new infections under the Omniscient baseline that observes the contact network, integrated across all budgets. Table \ref{tab:regret} shows the Total Regret for each approach on the two intervention problems. We also show the expected number of new infections at each budget in Figure \ref{fig:core-experiment}. Amongst predictive approaches, we saw average edit distances of 3748.0, 1721.6, and 1290.2 for the Pairwise Risk Classification, R2C-Diff, and Identity contact network predictions. 

\begin{table}[t!]
\centering
\begin{tabular}{@{}lcccc@{}}
\toprule
Approach & Intensive Treatment & Targeted PrEP\\
\midrule
No Intervention              & 373.0 & 405.7 \\
Random                       & 269.1 & 130.8 \\
Highest Degree               & 119.2 & 65.3 \\
PRC & 264.8 & 76.5 \\
Identity                     & 121.8 & 50.3\\
R2C-Agent (Ours)             & \textbf{53.8} & \textbf{35.8} \\
\emph{Omniscient} & \emph{0} & \emph{0}\\
\bottomrule
\end{tabular}
\caption{Total Regret for each approach on the two intervention problems. In each column, the best is \textbf{bolded} and Omniscient is italicized. PRC is Pairwise Risk Classification.}
\label{tab:regret}
\end{table}
 
First, we observe that R2C-Agent outperforms all baselines on intervention decisions, with a Total Regret 54.9\% lower on Intensive Treatment and 28.8\% lower on Targeted PrEP than the best baselines (Highest Degree and Identity, respectively), despite performing worse than Identity on average edit distance by a factor of 1.33. This empirically supports our earlier analysis from Section \ref{sec:generative} on how generative models can support decision making well despite having lower prediction accuracy. Second, we observe that R2C-Agent tends to have the greatest regret at extreme intervention budgets around 0.1-0.2, suggesting that errors in the predicted contact network are most costly when resources are most limited and each allocation carries more weight. Finally, we observe that R2C-Agent cuts the Total Regret of Random by factors of 5.00 and 3.65 on the Intensive Treatment and Targeted PrEP problems respectively, highlighting the potential value of network-based intervention strategies to make the most of limited resources.

\begin{figure}[t!]
    \centering
    \includegraphics[width=\linewidth]{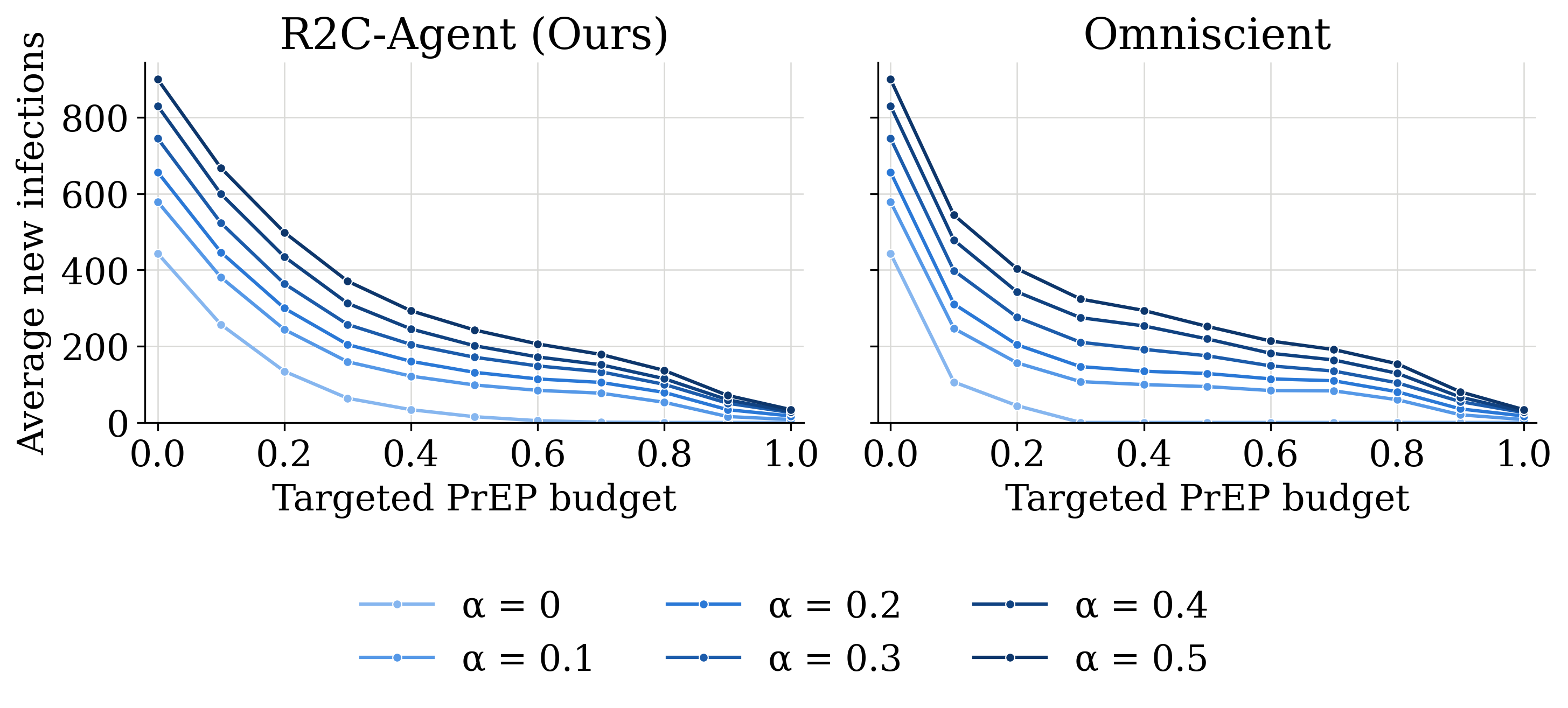}
    \caption{Performance ($\pm$ stratified SE) of R2C-Agent (left) and the Omniscient baseline (right) on the Targeted PrEP problem across different values of $\alpha$.}
    \label{fig:ablation1}
\end{figure}

For our first ablation study, we show how performance degrades as we violate the key assumption of our peer referral simulation, i.e., that nodes recruit along all of their edges. 
We naively plan interventions as before, but when we evaluate decisions, we modify the true test contact network prior to calculating spread by adding synthetic edges.
To add a synthetic edge, we first select a random node $v$ from the true contact network. 
Then, with probability 0.5, we add an edge between $v$ and a new synthetic node that did not participate in the peer referral network, which joins $V^{+}$ with probability $\frac{|V^{+}|}{|V|}$, and $V^{-}$ otherwise.
Otherwise, the new edge gets added between $v$ and a randomly selected true node from anywhere in $G$, including nodes in other components, that $v$ is not already connected to. 
The number of synthetic edges that we add is controlled with a corruption parameter $\alpha = \frac{\# \text{ of synthetic edges}}{\# \text{ of edges originally in } G_c}$. 
Figure \ref{fig:ablation1} shows how performance of R2C-Agent, as well as the Omniscient baseline, degrades on the Targeted PrEP problem as we vary $\alpha$ from 0 to 0.5.

First, we observe that R2C-Agent degrades about as gracefully as Omniscient, as Total Regret is about 28.3, 31.0, 25.8, 18.2, and 20.8 as $\alpha$ ranges from 0.1 to 0.5 (calculated against Omniscient at matching $\alpha$ values). Second, we note that overall spread increases substantially as additional positive and negative nodes are added to $V$, but the effect of targeting PrEP interventions via R2C-Agent qualitatively follows about the same trend and remains strongly preferable to a Random (untargeted) intervention. For example, even at the extreme $\alpha = 0.5$, the Total Regret of Random is 76.2 (compared to Omniscient at $\alpha = 0.5$), which is greater than that of R2C-Agent by a factor of 3.66.

\begin{figure}
    \centering
    \includegraphics[width=0.5\linewidth]{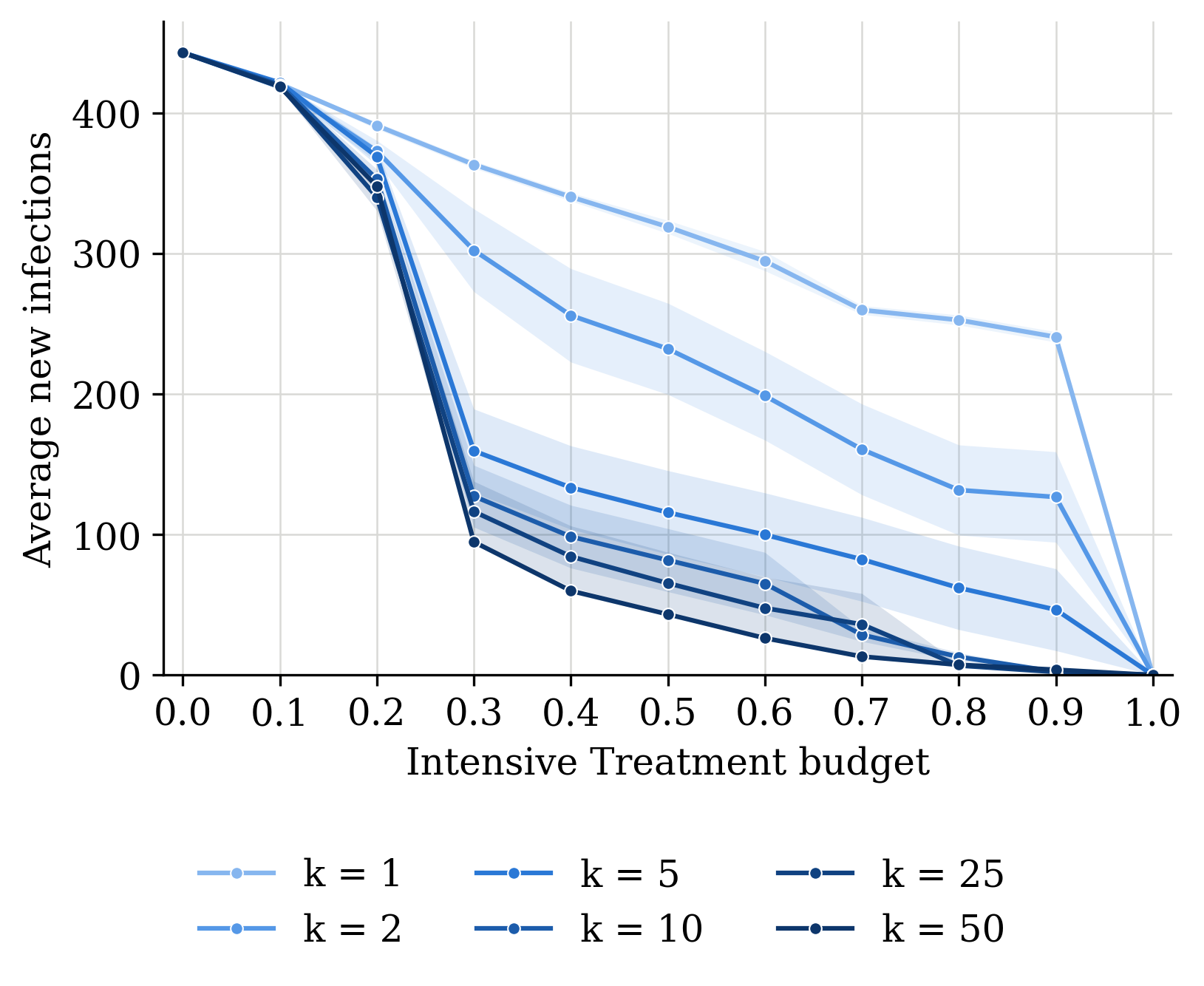}
    \caption{Performance ($\pm$ stratified SE) of R2C-Agent on the Intensive Treatment problem across different values of $k$.}
    \label{fig:ablation2}
\end{figure}

For our second ablation study, we demonstrate how the performance of R2C-Agent varies as we reduce the number of sampled contact networks that we explore but increase exploitation of each one. In all previous experiments, we set $k = 50$ contact network samples provided by R2C-Diff, and drew 1 Monte Carlo sample of HIV spread from each one to support the CAST and SandIMIN algorithms. Here, we fix the Monte Carlo sample budget at 50 but vary $k$, allowing us to draw $\frac{50}{k}$ Monte Carlo samples of HIV spread from each sampled contact network. Figure \ref{fig:ablation2} shows the results for R2C-Agent on the Intensive Treatment problem.

These results show that R2C-Agent favors exploration of different network samples over exploitation of a smaller number of network samples. Total Regret progresses from 53.8 at $k = 50$ to 101.3 at $k = 5$ to 240.4 at $k = 1$. This suggests that it is more important to plan over uncertainty in the network structure itself than over uncertainty in spread outcomes within fixed networks.

\section{Conclusion}\label{sec:conclusion}
Though intervention algorithms exist for strategically distributing limited HIV treatment and prevention resources, they require observability of the underlying contact network, which is prohibitively expensive to collect at scale. In this work, we study how to make intervention decisions when only the more scalable peer referral network is observable. We implement R2C-Agent, a decision making agent that uses R2C-Diff to infer distributions over possible contact networks conditioned on the corresponding peer referral network. We evaluate R2C-Agent's decisions on the ground truth contact network and show it outperforms baseline approaches from interdisciplinary literature on intervention effectiveness. Our results show that R2C-Agent has strong potential to support scalable network-based interventions for HIV treatment and prevention.

\paragraph{Path to Deployment.}
Datasets like the ICPSR dataset where both the contact network and the social network in a population are recorded are fundamental for potential future deployment of this research. This is because they allow us to extract pairs of contact networks and simulated peer referral networks for training R2C-Diff so that R2C-Agent can be deployed directly on an existing peer referral network project. Nevertheless, during a deployment, we would ideally want to label a small contact network within the peer referral network to act as an additional datapoint and to calibrate the peer referral simulation and R2C-Diff model to the population at hand so that we can extract better-aligned training data from the ICPSR dataset. See Appendix \ref{sec:ethics} for ethical considerations.

\paragraph{Future Work.}
In future work, we aim to study decision-focused variants of R2C-Diff. In particular, we are interested in whether we are able to directly optimize for minimizing downstream decision regret, and whether the learned model will generalize well to unseen networks when the training data volume is so limited. Another direction for future work is analyzing whether egocentric data can help support contact network sampling. For example, if we ask participants about their degree in the contact network without asking who their contacts are, how can that inform our model, and how robust can our model be to inaccurate responses? Finally, we will continue to evaluate our approach on other datasets and intervention applications and consider possible deployment avenues with existing peer referral network projects. 

\bibliographystyle{alpha}
\bibliography{sample}

@misc{unaids2026,
    author = {UNAIDS},
    title = {{Global AIDS Brief - United to end AIDS}},
    year = {2026},
    note = {Available at https://www.unaids.org/en/resources/documents/2026/20260612\_Global\_AIDS\_brief. Accessed September 29th, 2026}
}

@misc{unaids95,
    author = {UNAIDS},
    title = {{The Global AIDS Strategy for 2026-2031: United towards ending AIDS}},
    year = {2026}, 
    note = {Available at {https://www.unaids.org/en/2026-2031-global-aids-strategy}. Accessed September 29th, 2026}
}

@article{ten2025impact,
  title={Impact of an international HIV funding crisis on HIV infections and mortality in low-income and middle-income countries: a modelling study},
  author={Ten Brink, Debra and Martin-Hughes, Rowan and Bowring, Anna L and Wulan, Nisaa and Burke, Kelvin and Tidhar, Tom and Dalal, Shona and Scott, Nick},
  journal={The Lancet HIV},
  volume={12},
  number={5},
  pages={e346--e354},
  year={2025},
  publisher={Elsevier}
}

@article{khanna2019modeling,
  title={A modeling framework to inform PrEP initiation and retention scale-up in the context of Getting to Zero Initiatives},
  author={Khanna, Aditya S and Schneider, John A and Collier, Nicholson and Ozik, Jonathan and Issema, Rodal and Di Paola, Angela and Skwara, Abigail and Ramachandran, Arthi and Webb, Jeannette and Brewer, Russell and others},
  journal={AIDS (london, England)},
  volume={33},
  number={12},
  pages={1911},
  year={2019}
}

@article{kangaslahti2026policy,
  title={Policy-Embedded Graph Expansion: Networked HIV Testing with Diffusion-Driven Network Samples},
  author={Kangaslahti, Akseli and Choo, Davin and Kong, Lingkai and Tambe, Milind and van Heerden, Alastair and Johnson, Cheryl},
  journal={International Joint Conference on Artificial Intelligence (IJCAI)},
  year={2026}
}

@article{choo2026adaptive,
  title={Adaptive frontier exploration on graphs with applications to network-based disease testing},
  author={Choo, XianJun and Pan, Yuqi and Wang, Tonghan and Tambe, Milind and van Heerden, Alastair and Johnson, Cheryl and others},
  journal={Advances in Neural Information Processing Systems},
  volume={38},
  pages={149625--149666},
  year={2026}
}

@article{kong2026latent,
  title={Latent Spherical Flow Policy for Reinforcement Learning with Combinatorial Actions},
  author={Kong, Lingkai and Satish, Anagha and Jiang, Hezi and Kangaslahti, Akseli and Ma, Andrew and Chen, Wenbo and Song, Mingxiao and Xu, Lily and Tambe, Milind},
  journal={International Conference on Machine Learning (ICML)},
  year={2026}
}

@article{kangaslahti2026network,
  title={Network-Based Interventions for HIV Prevention via Cascade-Aware Suppression of Transmission},
  author={Kangaslahti, Akseli and Choo, Davin and Tambe, Milind and van Heerden, Alastair and Johnson, Cheryl},
  journal={arXiv preprint arXiv:2605.20218 (under review)},
  year={2026}
}

@article{mcknight2006respondent,
  title={Respondent-driven sampling in a study of drug users in New York City: notes from the field},
  author={McKnight, Courtney and Des Jarlais, Don and Bramson, Heidi and Tower, Lisa and Abdul-Quader, Abu S and Nemeth, Chris and Heckathorn, Douglas},
  journal={Journal of Urban Health},
  volume={83},
  number={Suppl 1},
  pages={54--59},
  year={2006},
  publisher={Springer}
}

@inproceedings{chen2016seeing,
  title={Seeing the unseen network: Inferring hidden social ties from respondent-driven sampling},
  author={Chen, Lin and Crawford, Forrest and Karbasi, Amin},
  booktitle={Proceedings of the AAAI Conference on Artificial Intelligence},
  year={2016}
}

@article{clipman2026geometric,
  title={Geometric deep learning enables high-fidelity network imputation for HIV transmission modeling},
  author={Clipman, Steven J and Wang, Jasmine and Mehta, Shruti H and Mohapatra, Shobha and Srikrishnan, Aylur K and Kumar, Muniratnam Suresh and Lucas, Gregory M and Latkin, Carl A and Solomon, Sunil S},
  journal={Communications Medicine},
  year={2026},
  publisher={Nature Publishing Group UK London}
}

@inproceedings{gilmer2017neural,
  title={{Neural message passing for quantum chemistry}},
  author={Gilmer, Justin and Schoenholz, Samuel S and Riley, Patrick F and Vinyals, Oriol and Dahl, George E},
  booktitle={International Conference on Machine Learning (ICML)},
  pages={1263--1272},
  year={2017}
}

@article{roberts2022impact,
  title={The impact of prevention-effective PrEP use on HIV incidence: a mathematical modelling study},
  author={Roberts, D Allen and Bridenbecker, Daniel and Haberer, Jessica E and Barnabas, Ruanne V and Akullian, Adam},
  journal={Journal of the International AIDS Society},
  volume={25},
  number={11},
  pages={e26034},
  year={2022},
  publisher={Wiley Online Library}
}

@article{liu2021efficient,
  title={Efficient network immunization under limited knowledge},
  author={Liu, Yangyang and Sanhedrai, Hillel and Dong, GaoGao and Shekhtman, Louis M and Wang, Fan and Buldyrev, Sergey V and Havlin, Shlomo},
  journal={National Science Review},
  volume={8},
  number={1},
  pages={nwaa229},
  year={2021},
  publisher={Oxford University Press}
}

@article{brazia2026reconstructing,
  title={Reconstructing MSM Sexual Networks to Guide PrEP Distribution Strategies for HIV Prevention},
  author={Br{\'a}zia, Jo{\~a}o and Kiss, Istv{\'a}n Z and Francisco, Alexandre P and Teixeira, Andreia Sofia},
  journal={arXiv preprint arXiv:2601.04434},
  year={2026}
}

@article{marcus2019use,
  title={Use of electronic health record data and machine learning to identify candidates for HIV pre-exposure prophylaxis: a modelling study},
  author={Marcus, Julia L and Hurley, Leo B and Krakower, Douglas S and Alexeeff, Stacey and Silverberg, Michael J and Volk, Jonathan E},
  journal={The Lancet HIV},
  volume={6},
  number={10},
  pages={e688--e695},
  year={2019},
  publisher={Elsevier}
}

@article{haas2021machine,
  title={Machine learning-based HIV risk estimation using incidence rate ratios},
  author={Haas, Oliver and Maier, Andreas and Rothgang, Eva},
  journal={Frontiers in Reproductive Health},
  volume={3},
  pages={756405},
  year={2021},
  publisher={Frontiers Media SA}
}

@inproceedings{yadav2016using,
  title={{Using Social Networks to Aid Homeless Shelters: Dynamic Influence Maximization under Uncertainty}},
  author={Yadav, Amulya and Chan, Hau and Jiang, Albert and Xu, Haifeng and Rice, Eric and Tambe, Milind},
  booktitle={International Conference on Autonomous Agents and Multiagent Systems (AAMAS)},
  year={2016}
}

@inproceedings{yadav2017influence,
  title={{Influence Maximization in the Field: The Arduous Journey from Emerging to Deployed Application}},
  author={Yadav, Amulya and Wilder, Bryan and Rice, Eric and Petering, Robin and Craddock, Jaih and Yoshioka-Maxwell, Amanda and Hemler, Mary and Onasch-Vera, Laura and Tambe, Milind and Woo, Darlene},
  booktitle={International Conference on Autonomous Agents and Multiagent Systems (AAMAS)},
  year={2017}
}

@inproceedings{wilder2018end,
  title={{End-to-End Influence Maximization in the Field}},
  author={Wilder, Bryan and Onasch-Vera, Laura and Hudson, Juliana and Luna, Jose and Wilson, Nicole and Petering, Robin and Woo, Darlene and Tambe, Milind and Rice, Eric},
  booktitle={International Conference on Autonomous Agents and Multiagent Systems (AAMAS)},
  year={2018}
}

@inproceedings{yadav2018please,
  title={{Please be an Influencer? Contingency-Aware Influence Maximization}},
  author={Yadav, Amulya and Noothigattu, Ritesh and Rice, Eric and Onasch-Vera, Laura and Marcolino, Leandro Soriano and Tambe, Milind},
  booktitle={International Conference on Autonomous Agents and Multiagent Systems (AAMAS)},
  year={2018}
}

@article{chen2023efficient,
  title={Efficient and degree-guided graph generation via discrete diffusion modeling},
  author={Chen, Xiaohui and He, Jiaxing and Han, Xu and Liu, Li-Ping},
  journal={International Conference on Machine Learning (ICML)},
  year={2023}
}

@article{vignac2022digress,
  title={Digress: Discrete denoising diffusion for graph generation},
  author={Vignac, Clement and Krawczuk, Igor and Siraudin, Antoine and Wang, Bohan and Cevher, Volkan and Frossard, Pascal},
  journal={International Conference on Learning Representations (ICLR)},
  year={2022}
}

@article{xu2024discrete,
  title={Discrete-state continuous-time diffusion for graph generation},
  author={Xu, Zhe and Qiu, Ruizhong and Chen, Yuzhong and Chen, Huiyuan and Fan, Xiran and Pan, Menghai and Zeng, Zhichen and Das, Mahashweta and Tong, Hanghang},
  journal={Advances in Neural Information Processing Systems},
  volume={37},
  pages={79704--79740},
  year={2024}
}

@article{liu2024graph,
  title={Graph diffusion transformers for multi-conditional molecular generation},
  author={Liu, Gang and Xu, Jiaxin and Luo, Tengfei and Jiang, Meng},
  journal={Advances in Neural Information Processing Systems},
  volume={37},
  pages={8065--8092},
  year={2024}
}

@article{jin2018learning,
  title={Learning multimodal graph-to-graph translation for molecular optimization},
  author={Jin, Wengong and Yang, Kevin and Barzilay, Regina and Jaakkola, Tommi},
  journal={International Conference on Learning Representations (ICLR)},
  year={2019}
}

@article{igashov2024equivariant,
  title={Equivariant 3D-conditional diffusion model for molecular linker design},
  author={Igashov, Ilia and St{\"a}rk, Hannes and Vignac, Cl{\'e}ment and Schneuing, Arne and Satorras, Victor Garcia and Frossard, Pascal and Welling, Max and Bronstein, Michael and Correia, Bruno},
  journal={Nature Machine Intelligence},
  volume={6},
  number={4},
  pages={417--427},
  year={2024},
  publisher={Nature Publishing Group UK London}
}

@inproceedings{kong2023autoregressive,
  title={Autoregressive diffusion model for graph generation},
  author={Kong, Lingkai and Cui, Jiaming and Sun, Haotian and Zhuang, Yuchen and Prakash, B Aditya and Zhang, Chao},
  booktitle={International conference on machine learning},
  pages={17391--17408},
  year={2023},
  organization={PMLR}
}

@misc{ICPSR,
    title = {{HIV Transmission Network Metastudy Project: An Archive of Data From Eight Network Studies, 1988--2001 (ICPSR 22140)}},
    author = {Martina Morris and Richard Rothenberg},
    year = {2011},
    howpublished = {ICPSR Data Holdings},
    note = {Available at doi.org/10.3886/ICPSR22140.v1}
}

@inproceedings{wang2024efficient,
  title={{Efficient Influence Minimization via Node Blocking}},
  author={Wang, Jinghao and Wu, Yanping and Wang, Xiaoyang and Zhang, Ying and Qin, Lu and Zhang, Wenjie and Lin, Xuemin},
  booktitle={Very Large Data Bases (VLDB)},
  year={2024}
}

@inproceedings{kempe2003maximizing,
  title={Maximizing the spread of influence through a social network},
  author={Kempe, David and Kleinberg, Jon and Tardos, {\'E}va},
  booktitle={Proceedings of the ninth ACM SIGKDD international conference on Knowledge discovery and data mining},
  pages={137--146},
  year={2003}
}

@article{austin2021structured,
  title={Structured denoising diffusion models in discrete state-spaces},
  author={Austin, Jacob and Johnson, Daniel D and Ho, Jonathan and Tarlow, Daniel and Van Den Berg, Rianne},
  journal={Advances in neural information processing systems},
  volume={34},
  pages={17981--17993},
  year={2021}
}

@inproceedings{sambaturu2020designing,
  title={Designing effective and practical interventions to contain epidemics},
  author={Sambaturu, Prathyush and Adhikari, Bijaya and Prakash, B Aditya and Venkatramanan, Srinivasan and Vullikanti, Anil},
  booktitle={Proceedings of the 19th International Conference on Autonomous Agents and MultiAgent Systems},
  year={2020}
}

@article{heckathorn1997respondent,
  title={Respondent-driven sampling: a new approach to the study of hidden populations},
  author={Heckathorn, Douglas D},
  journal={Social problems},
  volume={44},
  number={2},
  pages={174--199},
  year={1997},
  publisher={Oxford University Press Oxford, UK}
}

@article{bengtsson2012implementation,
  title={Implementation of web-based respondent-driven sampling among men who have sex with men in Vietnam},
  author={Bengtsson, Linus and Lu, Xin and Nguyen, Quoc Cuong and Camitz, Martin and Hoang, Nguyen Le and Nguyen, Tuan Anh and Liljeros, Fredrik and Thorson, Anna},
  journal={PLOS One},
  volume={7},
  number={11},
  pages={e49417},
  year={2012},
  publisher={Public Library of Science San Francisco, USA}
}

@article{stromdahl2015implementation,
  title={Implementation of web-based respondent driven sampling among men who have sex with men in Sweden},
  author={Str{\"o}mdahl, Susanne and Lu, Xin and Bengtsson, Linus and Liljeros, Fredrik and Thorson, Anna},
  journal={PLOS One},
  volume={10},
  number={10},
  pages={e0138599},
  year={2015},
  publisher={Public Library of Science San Francisco, CA USA}
}

@inproceedings{xia2014comparison,
  title={Comparison and validation of synthetic social contact networks for epidemic modeling},
  author={Xia, Huadong and Chen, Jiangzhuo and Marathe, Madhav V and Swarup, Samarth},
  booktitle={International Conference on Autonomous Agents and Multiagent Systems (AAMAS)},
  pages={1575--1576},
  year={2014}
}

@inproceedings{singh2018behavior,
  title={Behavior model calibration for epidemic simulations},
  author={Singh, Meghendra and Marathe, Achla and Marathe, Madhav V and Swarup, Samarth},
  booktitle={International Conference on Autonomous Agents and Multiagent Systems (AAMAS)},
  volume={2018},
  pages={1640},
  year={2018}
}

@inproceedings{yan2019maximization,
  title={On the Maximization of Influence Over an Unknown Social Network},
  author={Yan, Bo and Song, Kexiu and Liu, Jiamou and Meng, Fanku and Liu, Yiping and Su, Hongyi and others},
  booktitle={International Conference on Autonomous Agents and Multiagent Systems (AAMAS)},
  volume={19},
  pages={13--17},
  year={2019}
}

@article{kamarthi2019influence,
  title={Influence maximization in unknown social networks: Learning policies for effective graph sampling},
  author={Kamarthi, Harshavardhan and Vijayan, Priyesh and Wilder, Bryan and Ravindran, Balaraman and Tambe, Milind},
  journal={International Conference on Autonomous Agents and Multiagent Systems (AAMAS)},
  year={2020}
}

@inproceedings{godziszewski2023adversarial,
  title={Adversarial Link Prediction in Spatial Networks},
  author={Godziszewski, M and Vorobeychik, Yevgeniy and Michalak, Tomasz},
  booktitle={International Conference on Autonomous Agents and Multiagent Systems (AAMAS)},
  year={2023}
}

@article{li2020deep,
  title={Deep implicit coordination graphs for multi-agent reinforcement learning},
  author={Li, Sheng and Gupta, Jayesh K and Morales, Peter and Allen, Ross and Kochenderfer, Mykel J},
  journal={International Conference on Autonomous Agents and Multiagent Systems (AAMAS)},
  year={2021}
}

@article{amirkhanian2015effects,
  title={Effects of a social network HIV/STD prevention intervention for men who have sex with men in Russia and Hungary: a randomized controlled trial},
  author={Amirkhanian, Yuri A and Kelly, Jeffrey A and Takacs, Judit and McAuliffe, Timothy L and Kuznetsova, Anna V and Toth, Tamas P and Mocsonaki, Laszlo and DiFranceisco, Wayne J and Meylakhs, Anastasia},
  journal={AIDS (London, England)},
  volume={29},
  number={5},
  pages={583},
  year={2015}
}

@article{kelly2020social,
  title={Social network intervention to increase pre-exposure prophylaxis (PrEP) awareness, interest, and use among African American men who have sex with men},
  author={Kelly, Jeffrey A and Amirkhanian, Yuri A and Walsh, Jennifer L and Brown, Kevin D and Quinn, Katherine G and Petroll, Andrew E and Pearson, Broderick M and Rosado, A Noel and Ertl, Thom},
  journal={AIDS care},
  volume={32},
  number={sup2},
  pages={40--46},
  year={2020},
  publisher={Taylor \& Francis}
}

@article{grant2010preexposure,
  title={Preexposure chemoprophylaxis for HIV prevention in men who have sex with men},
  author={Grant, Robert M and Lama, Javier R and Anderson, Peter L and McMahan, Vanessa and Liu, Albert Y and Vargas, Lorena and Goicochea, Pedro and Casap{\'\i}a, Mart{\'\i}n and Guanira-Carranza, Juan Vicente and Ramirez-Cardich, Maria E and others},
  journal={New England Journal of Medicine},
  volume={363},
  number={27},
  pages={2587--2599},
  year={2010},
  publisher={Mass Medical Soc}
}

@article{baeten2012antiretroviral,
  title={Antiretroviral prophylaxis for HIV prevention in heterosexual men and women},
  author={Baeten, Jared M and Donnell, Deborah and Ndase, Patrick and Mugo, Nelly R and Campbell, James D and Wangisi, Jonathan and Tappero, Jordan W and Bukusi, Elizabeth A and Cohen, Craig R and Katabira, Elly and others},
  journal={New England journal of medicine},
  volume={367},
  number={5},
  pages={399--410},
  year={2012},
  publisher={Mass Medical Soc}
}

\newpage

\appendix

\section{Implementation Details} \label{sec:implementation}
We provide additional implementation details for our R2C-Diff model and experiments. Our full codebase can be accessed in our supplementary materials. We will publish the codebase as a public GitHub repository in the camera-ready version of the paper.
\subsection{R2C-Diff}
Our implementation of R2C-Diff uses node and pair embeddings of size 32. We use 3 message-passing layers with alternating pair updates (from both endpoints) and node updates (pair-weighted neighbor messages), with residual connections and dropout 0.2. Readout uses a 2-layer MLP. The total parameter count for the model is about 29,500, which is intentionally small to curb overfitting.

Training uses an Adam optimizer with learning rate 1e-3 and weight decay 1e-4. We train for 15 epochs. Training on a Mac with an Apple M4 Pro and MPS takes less than 2 hours.

\subsection{Data}
We use the HIV network from the ICPSR dataset \cite{ICPSR}. This is the network of people tested for HIV, with all recorded ties between two tested people. Contact ties include sex and needle edges, and all others are social ties. Fold assignment is done by sorting components by size, largest first, then assigning each component to the fold with fewer people so far.

\subsection{Experiments}
The Pairwise Risk Classifier constructs a vector for each node that includes a broader set of one-hot encoded covariates for each node, along with each node's HIV status. It then takes the element-wise sum and product of these vectors as features for two input nodes, along with the log of the component size. It is trained on all intra-component pairs in the training fold.

See \cite{kangaslahti2026network} and \cite{wang2024efficient} for details on CAST and SandIMIN, and see our code appendix for our implementation. We use a fixed edge probability of $p = 0.75$ for all edges across all experiments. See \cite{kangaslahti2026network} for behavior of the CAST algorithm under different edge probability initializations, including misspecified initializations.

Table \ref{tab:extended-results} shows the recorded means and stratified standard errors for R2C-Agent at each budget point, which are shown visually in Figure \ref{fig:core-experiment}.
\begin{table}[t!]
\centering
\begin{tabular}{@{}lcccc@{}}
\toprule
Budget & Intensive Treatment & Targeted PrEP\\
\midrule
0.0 & 442.8 ± 0.0 & 442.8 ± 0.0 \\
0.1 & 418.5 ± 0.7 & 256.4 ± 3.6 \\
0.2 & 347.7 ± 3.7 & 134.1 ± 2.7 \\
0.3 & 94.7 ± 1.6 & 63.4 ± 2.4 \\
0.4 & 59.9 ± 1.8 & 33.5 ± 1.7 \\
0.5 & 43.2 ± 1.1 & 15.1 ± 1.4 \\
0.6 & 26.3 ± 1.1 & 4.7 ± 0.7 \\
0.7 & 13.2 ± 0.9 & 0.7 ± 0.2 \\
0.8 & 7.3 ± 1.4 & 0.0 ± 0.0 \\
0.9 & 3.8 ± 0.9 & 0.0 ± 0.0 \\
1.0 & 0.0 ± 0.0 & 0.0 ± 0.0 \\
\bottomrule
\end{tabular}
\caption{Exact means $\pm$ stratified SE for average new infections across all budget points for R2C-Agent.}
\label{tab:extended-results}
\end{table}

\section{Ethics Statement} \label{sec:ethics}
This work consists only of retrospective simulation studies on anonymized data. We envision R2C-Agent as a decision support tool for experienced domain experts rather than an unsupervised automated policy. One possible ethical consideration in network-based interventions is the equity-efficiency tradeoff: optimizing for minimal average infections can divert resources away from individuals in smaller, more isolated connected components in the network. Tradeoffs like this should ultimately be assessed by public health practitioners rather than an algorithm like R2C-Agent alone. Any future pilot studies and trials will prioritize data privacy for anonymized participants and will be conducted in close collaboration with our partners and with appropriate institutional review board (IRB) approval and oversight. 

\end{document}